\pdfoutput=1
\documentclass[11pt]{article}
\usepackage{coling2020}
\usepackage{times}
\usepackage{url}
\usepackage{latexsym}

\usepackage{graphicx}
\usepackage{amsmath}
\usepackage{amsthm}
\usepackage{amsfonts}
\usepackage{booktabs}
\usepackage{algorithm}
\usepackage{algorithmic}
\colingfinalcopy 

\title{Automatic Distractor Generation for Multiple Choice Questions in Standard Tests}

\author{Zhaopeng Qiu \\ Tencent Medical AI Lab \\ {\tt zhaopengqiu@tencent.com} \\\And 
Xian Wu \thanks{\ \ Corresponding author}
\\ Tencent Medical AI Lab \\ {\tt kevinxwu@tencent.com} \\\AND
Wei Fan \\ Tencent Medical AI Lab \\ {\tt davidwfan@tencent.com} }

\date{}

\begin{document}
\maketitle

\begin{abstract}
To assess the knowledge proficiency of a learner, multiple choice question is an efficient and widespread form in standard tests. However, the composition of the multiple choice question, especially the construction of distractors is quite challenging. The distractors are required to both incorrect and plausible enough to confuse the learners who did not master the knowledge. Currently, the distractors are generated by domain experts which are both expensive and time-consuming. This urges the emergence of automatic distractor generation, which can benefit various standard tests in a wide range of domains.
In this paper, we propose a qu{\bf E}stion and answer guided {\bf D}istractor {\bf GE}neration (EDGE) framework to automate distractor generation. EDGE consists of three major modules: (1) the {\em Reforming Question Module} and the {\em Reforming Passage Module} apply gate layers to guarantee the inherent incorrectness of the generated distractors; (2) the {\em Distractor Generator Module} applies attention mechanism to control the level of plausibility. Experimental results on a large-scale public dataset demonstrate that our model significantly outperforms existing models and achieves a new state-of-the-art.
\end{abstract}

\section{Introduction}

\blfootnote{
    %
    %
    %
    %
    %
    
    \hspace{-0.5cm}  
    This work is licensed under a Creative Commons 
    Attribution 4.0 International License.
    License details:
    \url{http://creativecommons.org/licenses/by/4.0/}.
}

\begin{figure}[h]
	\centering \includegraphics[width=.8\linewidth]{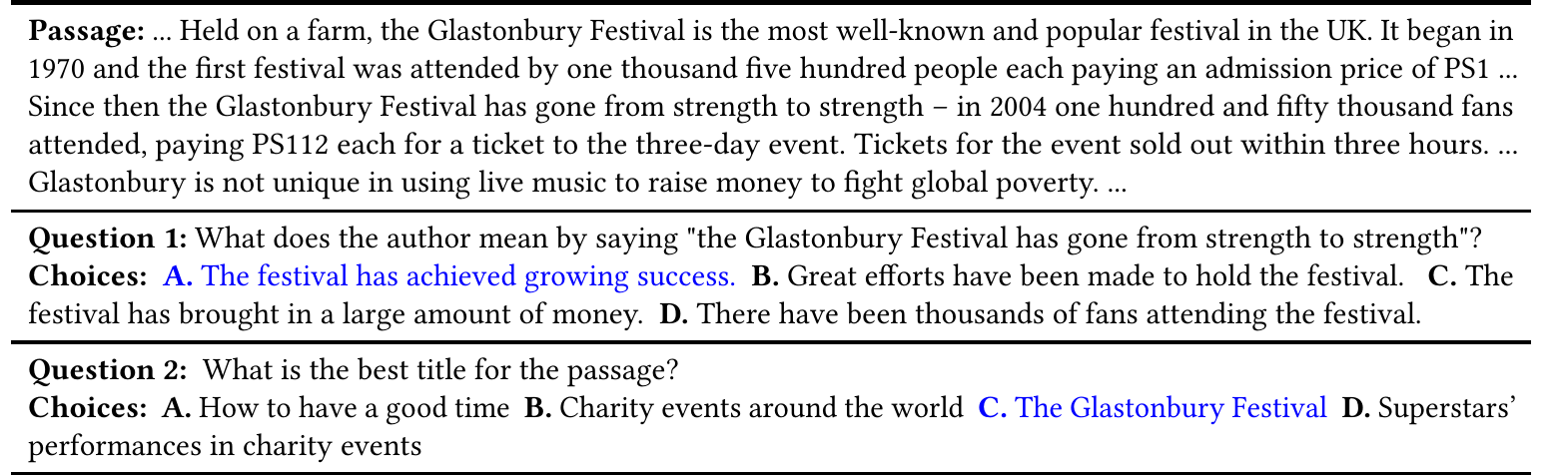}
	\caption{Two examples of multiple choice questions in RACE. Blue choices are the correct answers. \label{fig:question_example}}
\end{figure}

Standard test, such as TOFEL and SAT, is an efficient and essential tool to assess knowledge proficiency of a learner~\cite{Ch2018AutomaticMC}. According to testing results,
teachers or ITS (Intelligent Tutoring System) services can develop personalized study plans for different students. 
When organizing a standard test, a vital issue is to select a suitable question form. 
Among various question forms, multiple choice question (MCQ) is widely adopted in many notable tests, such as GRE, TOFEL and SAT.
MCQs have many advantages including less testing time, more objective and easy on the grader~\cite{Ch2018AutomaticMC}.
A typical MCQ consists of a stem and several candidate answers, among which one is correct, the rest are distractors.
As shown in Figure~\ref{fig:question_example}, in addition to a stem, some tests also include a long reading passage to provide the context of this MCQ.

The quality of an MCQ depends heavily on the quality of the distractors.
If the distractors can not confuse students, the correct answer could be concluded easily. As a result, the discrimination of the question will degrade, and the test will also lose the ability of the assessment.

However, it is a challenging job to design useful and qualified distractors.
Rather than being a trivial wrong answer, the distractor should have the plausibility which confuses learners who did not master the knowledge~\cite{Liang2018DistractorGF,DBLP:conf/cikm/QiuW019}.
A good distractor should be grammatically correct given the question and semantically consistent with the passage context of the question~\cite{Gao2018GeneratingDF}.
Meanwhile, the question composers need to enhance the plausibility of the distractor without hurting its inherent incorrectness.
Otherwise, the distractor easily becomes a definitely wrong answer, further making the question to be sloppy.
Hence, the manual preparation of distractors is time-consuming and costly~\cite{Ha2018AutomaticDS}.
It is an urgent issue to automatically generate useful distractors, which can help to alleviate question composers' workload and relax the restrictions on experience.
It could also be helpful to prepare a large train set to boost the machine reading comprehension (MRC) systems~\cite{Yang2017SemiSupervisedQW}. 
In this paper, we focus on automatically generating semantic-rich distractors for MCQs in real-world standard tests, such as RACE~\cite{Lai2017RACELR} which is collected from the English exams for Chinese students from grades 7 to 12.

Existing works conduct some attempts on generating short distractors~\cite{Stasaski2017MultipleCQ,Guo2016QuestimatorGK}.
These approaches formulate distractor generation as a similar word selection task.
They leverage the pre-defined ontology or word embeddings to find similar words/entities of the correct answer as the generated distractors.
These word-level or entity-level methods can only generate short distractors and do not apply to semantic-rich and long distractors for RACE-like MCQs.
Recently, generating longer distractors has been explored in a few studies~\cite{zhou2019coattention,Gao2018GeneratingDF}.
For example, ~\newcite{Gao2018GeneratingDF} proposes a sequence-to-sequence based model, which leverages the attention mechanism to automatically generate distractors from the reading passages. However, these methods mainly focus on the relation between the distractor and the passage and fail to comprehensively model the interactions among the passage, question and correct answer which helps to ensure the \textit{incorrectness} of the generated distractors.

To better generate useful distractors, we propose a novel qu\textbf{E}stion and answer guided \textbf{D}istractor \textbf{GE}neration (EDGE) framework.
More specifically, given the passage, the question and the correct answer, we first leverage a contextual encoder to generate the semantic representations for all text materials.
Then we use the attention mechanism to enrich the context of the question and the correct answer.
Next, we break down the distractor's usefulness into two aspects: the incorrectness and plausibility.
Incorrectness is the inherent attribute of the distractor, while plausibility refers to the ability to confuse the students.
We introduce two modules by leveraging the gate layer to guarantee the incorrectness: {\em Reforming Question Module} and {\em Reforming Passage Module}.
We further leverage an attention-based distractor generator, plus the previous two reforming modules, to guarantee the plausibility.
Finally, in the generation stage, we use the beam search to generate several diverse distractors by controlling their distances.
We conduct experiments on a large-scale public distractor generation dataset prepared from RACE. 
The experimental results demonstrate the effectiveness of our proposed framework.
Moreover, our method achieves a new state-of-the-art result in the distractor generation task.

\section{Related work}
In recent years, some research efforts have devoted to distractor generation. 
Generally, the related work can be classified into the following three categories.

\textbf{Feature-based methods}. 
Considerable research efforts~\cite{Liang2018DistractorGF,Sakaguchi2013DiscriminativeAT,Araki2016GeneratingQA} have been devoted to using manual design features, such as POS features and statistic features, to generate the distractors.
However, designing effective features is also labor intensive and hard to scale to various domains. Differently, our work is an end-to-end framework without manual design features.

\textbf{Similar word/entity-based methods}.
Some works~\cite{Stasaski2017MultipleCQ,Guo2016QuestimatorGK,kumar2015automatic,Afzal2014AutomaticGO} focused on finding answer-relevant ontologies or words as the distractors with the help of WordNet and Word2Vec.
For example, ~\newcite{Stasaski2017MultipleCQ} leveraged an educational Biology ontology to conduct the distractor generation.
However, some of these works depend heavily on the well-designed ontology and they can only generate short distractors, which usually only contain one single word or phrase. 

\textbf{NN-based methods.}
Recently, neural network based and data-driven solutions emerge.
%
\newcite{Gao2018GeneratingDF} proposed an end-to-end solution focusing on distractor generation for MCQs in standard English tests.
They employed the hierarchical encoder-decoder network as the base model and used the dynamic attention mechanism to generate the long distractors.
\newcite{zhou2019coattention} further strengthened the interaction between the question and the passage based on the model of \cite{Gao2018GeneratingDF}.


\section{Framework Description}

\subsection{Problem Definition}
In this paper, we focus on the automatic distractor generation for MCQs (see Figure~\ref{fig:question_example}). Let $P=\{w^p_t\}^{t=L_p}_{t=1}$ denote the reading passage, which consists of $L_p$ words. Let $Q=\{w^q_t\}^{t=L_q}_{t=1}$ and $A=\{w^a_t\}^{t=L_a}_{t=1}$ denote the question and its correct answer, respectively. $L_q$ and $L_a$ denote the lengths of the question and the answer, respectively. Note that the answer may not be a span of the passage $P$.

\paragraph{Problem Definition.} \textit{Formally, given the reading passage $P$, the question $Q$ and its correct answer $A$ as inputs, an EDGE model $\mathcal{M}$ aims to generate a distractor $D=\{w^d_t\}^{t=L_d}_{t=1}$ about the question, which is defined as finding the best distractor $\overline{D}$ that maximizes the conditional likelihood given $P$, $Q$, and $A$:}
\begin{equation*}
    \overline{D}=\underset{D}{\arg \max } \log \Pr(D | P, Q, A)
\end{equation*}

\subsection{Framework Overview}
\begin{figure*}
	\centering \includegraphics[width=.9\linewidth]{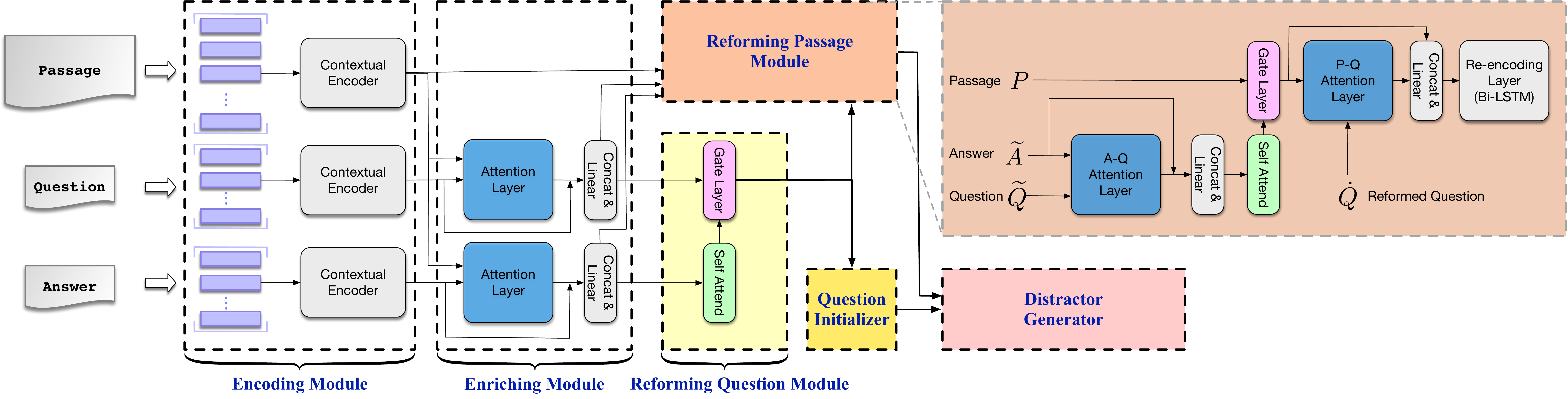}
	\caption{The EDGE framework.  \label{fig:framework}}
\end{figure*}

Inspired by existing question generation works~\cite{Duan2017QuestionGF,Du2017IdentifyingWT,Zhou2017NeuralQG,Kim2018ImprovingNQ}, we employ a sequence-to-sequence based network to generate the distractors.
As shown in Figure~\ref{fig:framework}, our overall framework contains five components. 
First, we employ the encoding module to extract the contextual semantic representations for all materials.
Then, we use the attention mechanism to enrich the semantic representations of the question and its answer.
Finally, we design three key components to generate useful distractors.

As mentioned above, the quality of the generated distractor are guaranteed from two aspects:
\begin{itemize}
    \item \textit{Incorrectness:} Both the passage and the question contain some parts strongly relevant to the answer, which may disorder the decoder to output the words contained by the answer and further hurt the inherent incorrectness of generated distractors. To guarantee the incorrectness, in our proposed framework, we reform the passage and question by erasing their answer-relevant information before they are fed into the decoder. Based on the gate mechanism, the two reforming modules highlight the distractor-relevant words and constrain the answer-relevant words by measuring the distances between the words and the correct answer.
    \item \textit{Plausibility:} To look reasonable, the distractor first should be grammatically and semantically consistent with the question. Otherwise, after reading the question, the students can trivially exclude it. Furthermore, to hinder students from excluding the distractor only by reading the passage, it should also be semantically relevant to the passage. To guarantee the plausibility, in our proposed framework, the distractor generator uses the semantic representation of the reformed question to initialize the generation process and leverages the attention mechanism to obtain the context representation from the reformed passage to guide the output.
\end{itemize}

We will address each component in detail in the following subsections.

\subsection{Encoding and Enriching Module}
In the encoding module, given a passage $P=\{w_t^p\}_{t=1}^{t=L_p}$, a question $Q=\{w_t^q\}_{t=1}^{t=L_q}$ and its answer $A=\{w_t^a\}_{t=1}^{t=L_a}$, we first convert every word $w$ to its $d$-dimensional vector $\mathbf{e}$ via an embedding matrix $\mathbf{E}\in \mathbb{R}^{|V|\times d}$, where $V$ is the vocabulary and $d$ is the dimension of word embedding. Then we use the encoder to extract the contextual representation for each word.
The outputs of the contextual encoder are three matrices: $\mathbf{P}\in \mathbb{R}^{L_p\times d}$, $\mathbf{Q}\in \mathbb{R}^{L_q\times d}$, and $\mathbf{A}\in \mathbb{R}^{L_a\times d}$. 



Next, we introduce the attention layer and the fusion layer to enrich the semantic representations of the question and its answer by fusing the passage information.
In the enriching module, we adopt the scaled dot product attention mechanism and the fusion kernel used in recent works~\cite{chen2017neural,mou2016natural} for better semantic understanding.
\begin{equation*}
    \mathbf{M}^{q} = \texttt{Attn}(\mathbf{Q}, \mathbf{P})=\textit{softmax}(\frac{\mathbf{Q}\mathbf{P}^\top}{\sqrt{d}}) 
\end{equation*}
\begin{equation*}
    \widetilde{\mathbf{Q}} = \texttt{Fuse}(\mathbf{Q}, \overline{\mathbf{Q}}) = \textit{tanh}([\mathbf{Q}; \overline{\mathbf{Q}}; \mathbf{Q} - \overline{\mathbf{Q}}; \mathbf{Q} \circ \overline{\mathbf{Q}}]\mathbf{W}_f+\mathbf{b}_f) \ \ \ \ \text{where  } \overline{\mathbf{Q}} = \mathbf{M}^{q} \mathbf{P} 
\end{equation*}
where $\mathbf{W}_f\in \mathbb{R}^{4d\times d}$ and $\mathbf{b}_f\in \mathbb{R}^d$ are the parameters to learn.
$[;]$ denotes the column-wise concatenation. $\circ$ and $-$ denote the element-wise multiplication and subtraction between two matrices, respectively. $\mathbf{M}^{q}\in \mathbb{R}^{L_q\times L_p}$ denotes the attention weight matrix.

For the answer $\mathbf{A}$, we can also obtain the enriched representation $\widetilde{\mathbf{A}}$ via the same attention and fusion process.

\subsection{Reforming Question Module}
As mentioned above, to retain the inherent incorrectness of the distractors, we need to reform the question by constraining the answer-relevant parts.
In this module, we first evaluate the semantic distance between the answer and each word of the question. 
Then, the distances are used as the weights to differentiate the useful words from the words strongly relevant to the correct answer.
The reforming process is conducted in the following two steps.

\textbf{Self-Attend Layer.} Firstly, we use this layer to obtain the sentence-level representation $\widetilde{\mathbf{v}}^a\in \mathbb{R}^{1\times d}$ of the answer.
\begin{equation*}
     \widetilde{\mathbf{v}}^a = \texttt{SelfAlign}(\widetilde{\mathbf{A}}) = \mathbf{r}^{\top} \widetilde{\mathbf{A}} \ \ \ \ \text{where  } \mathbf{r} = softmax(\widetilde{\mathbf{A}}\mathbf{W}_a) 
\end{equation*}
where $\textbf{W}_a\in \mathbb{R}^{d\times 1}$ is a trainable parameter.
$\mathbf{r}\in \mathbb{R}^{L_a\times 1}$ denotes the weight vector.

\textbf{Gate Layer.} In this layer, we first use a bilinear layer to measure the distance between each word and the answer.
Then, the distance information is used as gate values to reform each word and gain the reformed question $\dot{\mathbf{Q}}$.
\begin{equation*}
    \dot{\textbf{Q}}_{i} = \delta_i \widetilde{\textbf{Q}}_{i}, i\in[1, L_q]\ \ \ \ \text{where  } \delta_i = \texttt{Gate}(\widetilde{\textbf{Q}}_{i}, \widetilde{\mathbf{v}}^{a}) = \widetilde{\textbf{Q}}_{i} \textbf{W}_g^q \widetilde{\mathbf{v}}^{a\top} + b_g^q
\end{equation*}
where $\textbf{W}_g^q$ is a trainable bilinear projection matrix and $b_g^q \in \mathbb{R}$ is also a parameter to learn. $\delta_i\in \mathbb{R}$ is the semantic distance between the correct answer and the $i$-th word of the question. $\widetilde{\textbf{Q}}_{i}$ and $\dot{\textbf{Q}}_{i}$ denote the original representation and reformed representation of the $i$-th word in the question, respectively.

\subsection{Reforming Passage Module}
Likewise, the passage also needs to be reformed to erase the impact of the answer.
However, there still are some differences between the two reforming modules' architectures:
\begin{itemize}
    \item Some parts of the passage may belong to other questions but have some common words with the answer. These common words may obtain low gate values which further reduce their contributions to the generation. Hence, before passage reforming, we first attend the question information to the answer to constrain it only to affect the words related to its question.
    \item To further strengthen the relationship between the generated distractor and the question, we integrate the question information into the reformed passage to further highlight the question-relevant sentences.
\end{itemize}
The reformation process consists of the following four steps.

\textbf{A-Q Attention Layer.} We use the $\texttt{Attn}(\cdot, \cdot)$ and $\texttt{Fuse}(\cdot, \cdot)$ to fuse the question information into the answer. Note that we use the original question but not the reformed question because the former contains complete answer-relevant information.
\begin{equation*}
    \hat{\mathbf{A}} = \texttt{Fuse}(\widetilde{\mathbf{A}}, \overline{\mathbf{A}})\ \ \ \ \text{where  } \overline{\mathbf{A}} = \texttt{Attn}(\widetilde{\mathbf{A}}, \widetilde{\mathbf{Q}})\widetilde{\mathbf{Q}}
\end{equation*}

\textbf{Self-Attend} \& \textbf{Gate Layer.} As the reforming question module, we first use the $\texttt{SelfAlign}(\cdot)$ to obtain the answer's sentence-level representation $\hat{\mathbf{v}}^a=\texttt{SelfAlign}(\hat{\mathbf{A}})\in \mathbb{R}^{1\times d}$.
We then use another gate layer to obtain the reformed passage $\dot{\mathbf{P}}$.
\begin{equation*}
    \dot{\mathbf{P}}_{i} = \texttt{Gate}(\mathbf{P}_{i}, \hat{\mathbf{v}}^{a})\mathbf{P}_{i}, i\in[1, L_p]
\end{equation*}
where $\mathbf{P}_{i}$ and $\dot{\mathbf{P}}_i$ denote the original and reformed representations of the $i$-th passage word, respectively.

\textbf{P-Q Attention Layer.} This layer uses the attention mechanism to fuse the question information as mentioned above.
\begin{equation*}
    \widetilde{\mathbf{P}} = \texttt{Fuse}(\dot{\mathbf{P}}, \overline{\mathbf{P}})\ \ \ \ \text{where  }\overline{\mathbf{P}} = \texttt{Attn}(\dot{\mathbf{P}}, \dot{\mathbf{Q}})\dot{\mathbf{Q}}
\end{equation*}
    

\textbf{Re-encoding Layer.} We re-extract the contextual representation by another Bi-LSTM for the reformed passage $\widetilde{\textbf{P}}$. 
The final semantic representation of the passage is denoted as $\hat{\mathbf{P}}\in \mathbb{R}^{L_p\times d}$.

\subsection{Question Initializer} 
As mentioned above, the generated distractor should be grammatically and semantically consistent with the question. Inspired by \cite{Gao2018GeneratingDF}, we use the question information to initialize the decoding process to enhance the semantic relevance between the distractor and question. 
Specifically, we first use a Bi-LSTM~\cite{hochreiter1997long} to re-encode the reformed question $\dot{\textbf{Q}}$.
\begin{equation*}
    \overrightarrow{\mathbf{h}^q_i} = \overrightarrow{\textit{LSTM}}(\dot{\textbf{Q}}_{i},  \overrightarrow{\mathbf{h}^q_{i-1}})
\end{equation*}
where $\overrightarrow{\mathbf{h}^q_i}$ is the hidden state of the forward LSTM at time $i$.
We then concatenate the last hidden states of two directions as $\mathbf{h}^q\in \mathbb{R}^{d}$, which then is projected to get the initial state of the decoder $\mathbf{h}_0$.
\begin{equation*}
\mathbf{h}_0 = \mathbf{h}^q \mathbf{W}_p + \mathbf{b}_p\ \ \ \ \text{where  }\mathbf{h}^q = [\overrightarrow{\mathbf{h}^q_{L_q}};\overleftarrow{\mathbf{h}^q_{L_q}}]
\end{equation*}
where $\mathbf{W}_p\in \mathbb{R}^{d\times d}$ and $\mathbf{b}_p\in \mathbb{R}^{d}$ are learnable parameters.

\subsection{Distractor Generator}
At the decoder side, we adopt an attention-based LSTM layer.
Specifically, at the first step, we use the output of the reforming question module $\mathbf{h}_0$ as the initial state and use the mean pooling vector of the reformed passage $\hat{\mathbf{P}}$ as the context vector $\mathbf{c}_0\in \mathbb{R}^{d}$.
The first word is set to the special token \textit{[EOS]}.
\begin{equation*}
    \mathbf{c}_0 = \texttt{MeanPooling}(\hat{\mathbf{P}}),\ \ \  \mathbf{e}_0 = \mathbf{E}(\textit{[EOS]})
\end{equation*}
Next, for each decoding step $t$, we use the attention mechanism to attend the most relevant words in the reading passage to form the context vector.
\begin{equation}
    \label{eq:decode}
        \textbf{h}_t = \textit{LSTM}([\mathbf{e}_{t-1}; \mathbf{c}_{t-1}], \mathbf{h}_{t-1}), \ \ \ 
        \textbf{c}_t = \texttt{Attn}(\mathbf{h}_t \mathbf{W}_h, \hat{\mathbf{P}})  \hat{\mathbf{P}}
\end{equation}
where  $\textbf{W}_h \in \mathbb{R}^{d\times d}$, projecting the hidden state to the passage context, is the parameter to learn. $\mathbf{e}_{t-1}$ denotes the embedding of the word at the $t-1$-th step.

Moreover, at each step, we concatenate $\textbf{h}_t$ and $\textbf{c}_t$ together and use an MLP layer to predict the word probability distribution.
\begin{equation}
\label{eq:loss}
    H_V = \textit{softmax}(\textit{tanh}([\textbf{h}_t; \textbf{c}_t]\textbf{W}_s) \textbf{W}_v + \textbf{b}_v)
\end{equation}
where $\textbf{W}_s\in \mathbb{R}^{2d\times d}$, $\textbf{W}_v \in \mathbb{R}^{d\times |V|}$ and $\textbf{b}_v\in \mathbb{R}^{|V|}$ are learnable parameters. $H_V$ denotes the probabilities of all words in the vocabulary in which the word with the maximum probability is the predicted word at step $t$.

\subsection{Training and Inference}
We train the model by minimizing regular cross-entropy loss:
\begin{equation*}
    \mathcal{L}(\theta_{\mathcal{M}}) = -\sum_{\mathcal{D}}\sum_{t}\log \Pr (w_t^d|P,A,Q,w_{<t}^d;\theta_{\mathcal{M}})
\end{equation*}
where $\mathcal{D}$ is the training corpus in which each data sample contains a distractor $D$, a passage $P$, a question $Q$ and an answer $A$. $w_t^d$ is the $t$-th position of the distractor $D$. $\Pr (w_t^d|P,A,Q,w_{<t}^d;\theta)$ is the predicted probability of the $w_t^d$ and can be calculated by the Eq.(\ref{eq:loss}). $\theta_{\mathcal{M}}$ denotes all trainable parameters in EDGE.

During the inference phase, we use a beam search of width $n$ and receive $n$ candidate distractors with decreasing likelihood because an MCQ has several diverse distractors.
Following \newcite{Gao2018GeneratingDF}, we use the Jaccard distance to generate the final multiple diverse distractors from the beam search results.
Specifically, we first select the first candidate distractor with the maximum likelihood from the search results as $D_1^g$. 
The second one $D_2^g$ should have a Jaccard distance, larger than 0.5, to $D_1^g$.
Likewise, we select the third one $D_3^g$ which has a restricted distance to both $D_1^g$ and $D_2^g$.

\section{Experiments}

\subsection{Experiment Setup}
\subsubsection{Dataset}
For a fair comparison, we use the distractor generation dataset\footnote{https://github.com/Evan-Gao/Distractor-Generation-RACE} released by \newcite{Gao2018GeneratingDF} as our benchmark. 
This dataset is constructed based on RACE~\cite{Lai2017RACELR}, which is collected from the English exams and widely used in the MRC field. 
More details about the dataset construction process can be found in \cite{Gao2018GeneratingDF}.
The train/validation/test set contain 96,501/12,089/12,284 examples, respectively.

\begin{figure}[h]
\centering
\includegraphics[width=0.45\linewidth]{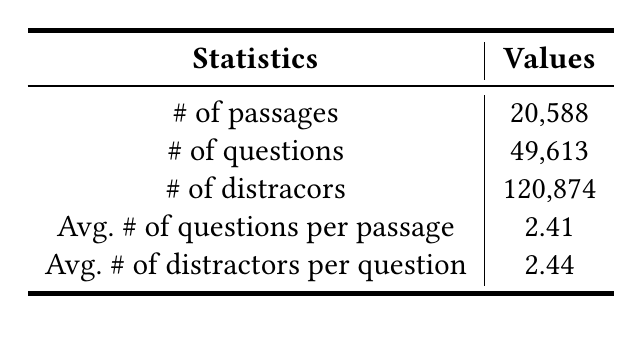}
\includegraphics[width=0.45\linewidth]{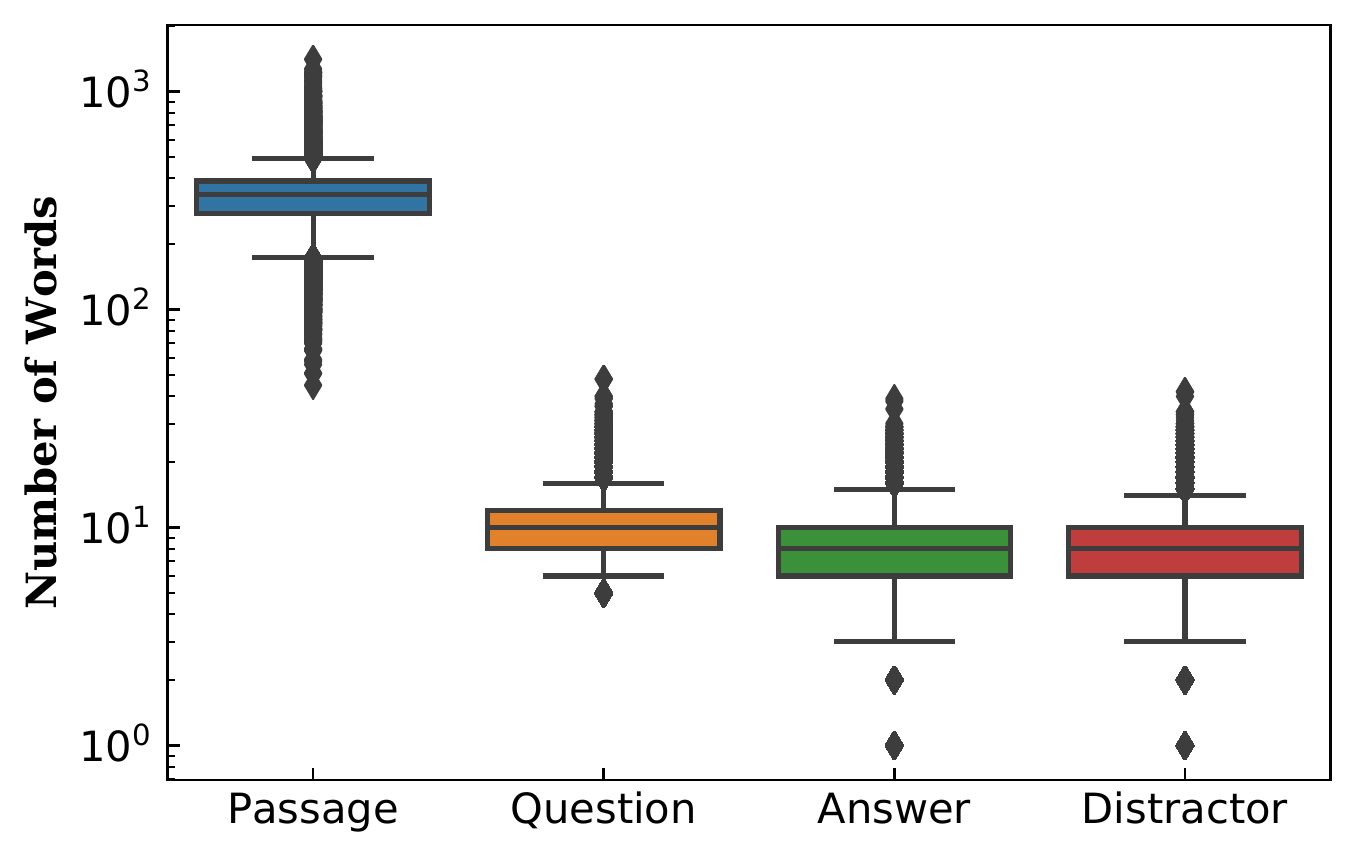}
\caption{The statistics of the evaluation dataset (Left). The word count distribution (Right). The max values are 95th percentile.  \label{fig:data_stats}}
\end{figure}

The left sub-figure of Figure~\ref{fig:data_stats} shows the count statistics on this dataset.
We can see that most passages are related to more than two questions.
A question is usually associated with multiple distractors, which proves necessary to conduct the beam search in the testing phase.
The right sub-figure shows the distributions of the lengths of the passages, questions, answers, and distractors.
The distractors and the answers have similar lengths and the questions are slightly longer than them.
Meanwhile, we can see that the median of the distractor lengths is larger than 8.
This suggests the similar word/entity-based methods do not apply to this dataset.

\subsubsection{Model Details}
We use the \textit{GloVe.840B.300d}~\cite{pennington2014glove} as the pre-trained word embeddings (i.e., $d=300$), and the word representations are shared across different components of EDGE.
In the encoding module, we choose the Bi-LSTM as the contextual encoder, the size of the hidden unit is set to 300 (150 for each direction).
Please note that the Bi-LSTM encoder is a plug-in module that can be easily replaced by Transformer~\cite{Vaswani2017AttentionIA}, BERT~\cite{Devlin2019BERTPO} or XLNet~\cite{Yang2019XLNetGA}.
The parameters of the Bi-LSTM are shared among the encoding module and two reforming modules.

According to the 95th percentile values shown in Figure~\ref{fig:data_stats}, we set the maximum lengths of passages, questions, answers, and distractors to be 500, 17, 15, and 15, respectively.

The model is trained with a mini-batch size of 64. 
We use Nesterov Accelerated Gradient (NAG) optimizer~\cite{Nesterov1983AMF} with a learning rate of 0.005. 
The dropout rate is set to 0.1 to reduce overfitting.
The beam size $n$ is set to 50.

\subsection{Baseline Approaches and Metrics}
The following models are selected as baselines:

\textbf{Basic models:} the basic sequence-to-sequence framework and its variants including (1) \textbf{SS} (Seq2Seq): the basic model that generates a distractor from the passage; (2) \textbf{SEQ} (SS+Enriching Module+Question Initializer): the sequence-to-sequence with the enriching module and the question-initialized decoder in which the initial state is set to the output of the question initializer; and (3) \textbf{SEQA} (SEQ+Attention): the sequence-to-sequence with a decoder same as the distractor generator of the EDGE.



\textbf{HRED (HieRarchical Encoder-Decoder):} the basic framework of \cite{Gao2018GeneratingDF}, which also contains the question initializer and attention mechanism.

\textbf{HSA (HRED+static attention)}~\cite{Gao2018GeneratingDF}: which uses the HRED as the basic architecture and leverages two attention strategies to combine the information of the passage, question, and answer.

\textbf{CHN}~\cite{zhou2019coattention}: which extends HSA with a co-attention mechanism to further strengthen the interaction between the passage and the question. This model achieved state-of-the-art performance previously on this task.

Following the distractor generation work~\cite{Gao2018GeneratingDF} and some question generation works~\cite{Kim2018ImprovingNQ,Chen2019ReinforcementLB}, we use BLEU~\cite{Papineni2001BleuAM} and ROUGE~\cite{Lin2004ROUGEAP} scores as our evaluation metrics.

All hyper-parameters of EDGE and other baselines are selected on the validation set based on the lowest perplexity and the results are reported on the test set.

\subsection{Performance Comparison}
\label{sec:perf}
The experimental results of all models are summarized in Table~\ref{tab:performance}.
Since the dataset in ~\cite{Gao2018GeneratingDF} is slightly different from the public dataset in Github, we not only report the HSA's results from its original paper ~\cite{Gao2018GeneratingDF} but also include the reimplementation results of \cite{Gao2018GeneratingDF} from ~\cite{zhou2019coattention} on the public dataset.
\begin{table*}[h]
\centering
\caption{The performance comparison results. The best results are highlighted bold. HSA* denotes the result reported in ~\cite{Gao2018GeneratingDF} and HSA denotes the result reported in \cite{zhou2019coattention}.  \label{tab:performance}}
\resizebox{.8\textwidth}{!}{
\begin{tabular}{l|c|c|c|c|c|c|c}
\toprule[1pt]
1st Distractor  & BLEU-1   &  BLEU-2  & BLEU-3   & BLEU-4    & ROUGE-1  & ROUGE-2    & ROUGE-L        \\ 
\midrule
SS      & 18.76 & 7.84 & 3.82 & 1.91  & 13.46   & 2.73 & 13.60  \\
SEQ  & 24.44 & 11.01 & 5.59 & 3.02  & 16.41    & 3.74  & 14.85   \\
SEQA & 28.61 & 15.28 & 9.47 & 6.3  & 18.92   & 5.56  & 15.53 \\ 
HRED & 27.96 & 14.41 & 9.05 & 6.34  & 14.12   & 3.97  & 14.68      \\
HSA* & 27.32 & 14.69 & 9.29 & 6.47  & 15.69   & 4.42  & 15.12      \\
HSA & 28.18 & 14.57 & 9.19 & 6.43 & 15.74 & 4.02 & 14.89 \\
CHN & 28.65 & 15.15 & 9.77 & 7.01 & 16.22 & 4.34 & 15.39 \\
\textbf{EDGE} & \textbf{33.03} & \textbf{18.12} & \textbf{11.35} & \textbf{7.57}  & \textbf{19.63}   & \textbf{5.81}  & \textbf{19.24}      \\
\midrule
\midrule
2nd Distractor  & BLEU-1   &  BLEU-2  & BLEU-3   & BLEU-4    & ROUGE-1  & ROUGE-2    & ROUGE-L        \\ 
\midrule
SS      & 18.19 & 7.15 & 3.24 & 1.51  & 13.12   & 2.35 & 13.33  \\
SEQ  & 24.18 & 10.48 & 5.05 & 2.63  & 15.99    & 3.31  & 14.36   \\
SEQA & 28.00 & 14.20 & 8.20 & 5.04  & 18.24   & 4.78  & 14.79 \\ 
HRED & 27.85 & 13.39 & 7.89 & 5.22  & 15.51   & 3.44  & 14.48      \\
HSA* & 26.56 & 13.14 & 7.58 & 4.85  & 14.72   & 3.52  & 14.15      \\
HSA & 27.85 & 13.41 & 7.87 & 5.17 & 15.35 & 3.40 & 14.41 \\
CHN & 27.29 & 13.57 & 8.19 & 5.51 & 15.82 & 3.76 & 14.85 \\
\textbf{EDGE} & \textbf{32.07} & \textbf{16.75} & \textbf{9.88} & \textbf{6.27}  & \textbf{18.53}   & \textbf{4.81}  & \textbf{18.10}      \\ 
\midrule
\midrule
3rd Distractor  & BLEU-1   &  BLEU-2  & BLEU-3   & BLEU-4    & ROUGE-1  & ROUGE-2    & ROUGE-L        \\ 
\midrule
SS      & 18.60 & 7.38 & 3.33 & 1.56  & 13.33   & 2.40 & 13.41  \\
SEQ  & 24.20 & 10.24 & 4.85 & 2.47  & 15.63    & 3.12  & 13.99   \\
SEQA & 27.26 & 13.60 & 7.73 & 4.72  & 17.61   & 4.19  & 15.18 \\ 
HRED & 26.73 & 12.55 & 7.21 & 4.58 & 15.96   & 3.46  & 14.86      \\
HSA* & 26.92 & 12.88 & 7.12 & 4.32  & 14.97   & 3.41  & 14.36      \\
HSA & 26.93 & 12.62 & 7.25 & 4.59 & 15.80 & 3.35 & 14.72 \\
CHN & 26.64 & 12.67 & 7.42 & 4.88 & 16.14 & 3.44 & 15.08 \\
\textbf{EDGE} & \textbf{31.29} & \textbf{15.94} & \textbf{9.24} & \textbf{5.70}  & \textbf{17.83}   & \textbf{4.40}  & \textbf{17.46}      \\ 
\bottomrule[1pt]
\end{tabular}
}
\end{table*}
There are several observations:
Firstly, the proposed model, EDGE, outperforms all baselines significantly in all metrics and achieves the new state-of-the-art scores on this distractor generation dataset;
Secondly, SEQ and SEQA outperform the basic Seq2Seq which indicates both the question information and the passage information are vital to the distractor generation;
Thirdly, SEQA outperforms HRED which indicates that the co-attention between the question and the passage in the enriching module can help to generate better distractors.
The observation that CHN outperforms HSA also proves the effectiveness of the co-attention mechanism;
Finally, the basic Seq2Seq performs far worse than other models, which indicates that the distractor generation is a challenging task and hard to solve only with simple models.


\subsection{Ablation Analysis}
\begin{table}
\centering
\caption{The ablation study results. We average the BLEU-4 and ROUHE-L over all three generated distractors. Higher scores indicate better performance. \label{tab:ablation}}
\resizebox{.55\columnwidth}{!}{
\begin{tabular}{l|c|c}
\toprule[1pt]
Methods  & BLEU-4  & ROUGE-L        \\ 
\midrule
EDGE & 6.51 & 18.27   \\
\midrule
w/o Reforming Passage Module & 5.72 & 17.12 \\
w/o Reforming Question Module & 6.01 & 17.84 \\
w/o Question Initializer & 6.12  & 18.01 \\
w/o Enriching Module & 6.41 &  18.12 \\
w/o Encoding Module & 1.35 & 9.07 \\
\bottomrule[1pt]
\end{tabular}
}
\end{table}
Table~\ref{tab:ablation} shows the experimental results of the ablation study.
We can see that removing the \textit{Reforming Passage Module} or the \textit{Reforming Question Module} leads to the suboptimal results.
This validates the effectiveness of two reforming mechanisms.
Moreover, we find the former module is more important for the overall distractor generation model.
This is probably due to that the reformed passage has a higher impact on the decoding process. Particularly, in each decoding step, the context information from the passage can provide more clues to generate proper distractors than the context information from the question.

We can also observe that the question initializer brings performance gain. This verifies the hypothesis in Section 3.6 that the initial decoder state encoded from the question helps to generate distractors grammatically and semantically consistent with the question.
Removing the encoding or enriching module will also result in a performance drop. This indicates that extracting contextual representations is important for the generation task.
Moreover, the enriching module can further improve the performance by fusing the passage information into the contextual representation of the question.

\subsection{Human Evaluation}
\begin{table}
\centering
\caption{Results of human evaluation. Higher scores indicate better performance. \label{tab:human}}
\resizebox{.5\columnwidth}{!}{
\begin{tabular}{l|c|c|c}
\toprule[1pt]
Methods  & Fluency  & Coherence & Distracting Ability        \\ 
\midrule
HSA & 7.84 & 5.46 & 3.51 \\
CHN & 8.19 & 5.80 & 4.70 \\
\textbf{EDGE} & \textbf{8.95} & \textbf{7.08} & \textbf{5.50}   \\
\bottomrule[1pt]
\end{tabular}
}
\end{table}
We conduct a human evaluation to evaluate the quality of the generated distractors of different models.
We use three metrics designed by ~\newcite{zhou2019coattention} to conduct the evaluation: (1) \textit{Fluency}, which evaluates whether the distractor follows regular English grammar and conforms with human logic and common sense; (2) \textit{Coherence}, which measures whether the key phrases in the distractors are relevant to the passage and the question; (3) \textit{Distracting Ability}, which evaluates how likely a generated distractor will be used by the question composers in real examinations.
We choose the first 100 samples of the test set and the corresponding distractors generated by three models as the input.
We employ five annotators with good English background (at least holding a bachelor's degree) to scores these distractors with three gears (i.e., \textit{Good}, \textit{Fair} or \textit{Bad}) by three metrics, the scores are then projected to 0 - 10. 

The results of all models, averaged over all generated distractors, are shown in Table~\ref{tab:human}. We can find that our model performs best in three metrics. This suggests our model is able to generate plausible and useful distractors. This conclusion also aligns with the experimental results of automatic metrics in Section \ref{sec:perf}.

\subsection{Case Study}

\begin{figure}
\centering
\includegraphics[width=\linewidth]{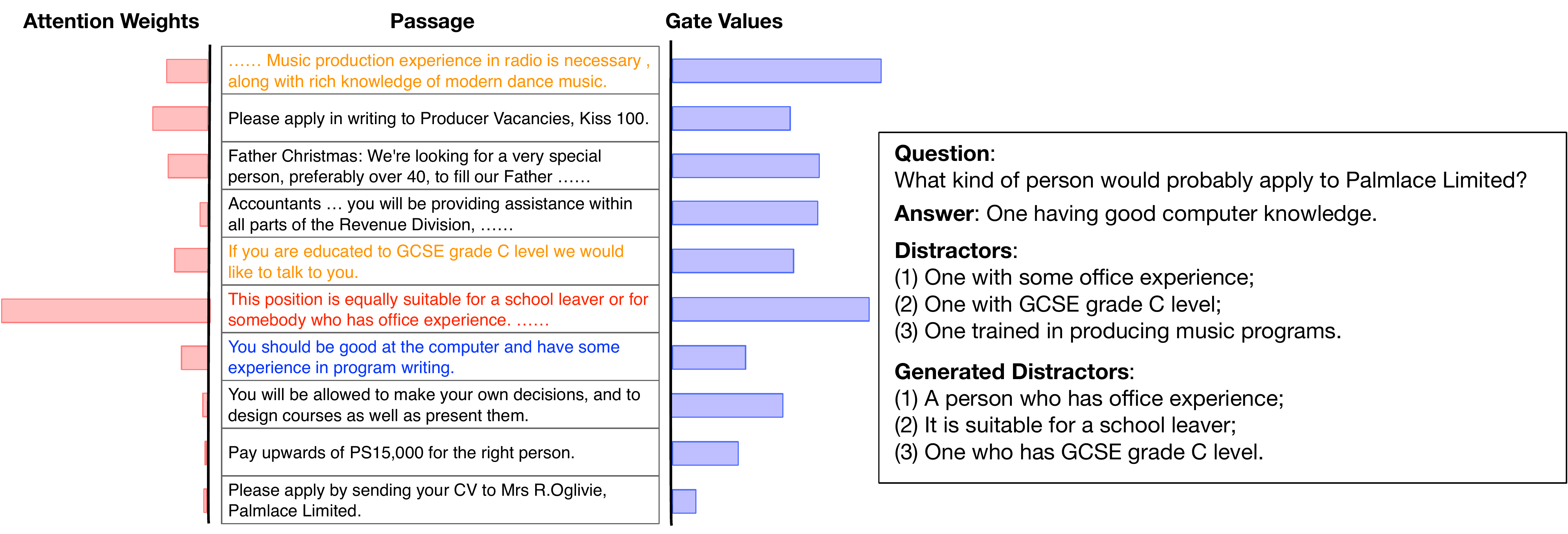}
\caption{A sample question with the truth distractors and generated distractors. The left sub-figure shows the attention weights in the generator and the gate values in the reforming passage module when decoding the first generated distractor. To enable comparisons among different sentences, we average the attention and gate values of all words in each sentence. Colored sentences are the clues of four options.  \label{fig:case}}
\end{figure}
The design of two reforming modules and the generator enables convenient interpretation of the generated distractors.
Take the MCQ in Figure~\ref{fig:case} for example, the \textit{blue} highlighted sentence in the paragraph includes the clue to infer the correct answer. EDGE managed to block this clue by assigning a lower attention weight with the help of the gate layer. In this manner, the clue of the correct answer is prohibited from generating the distractor.
Meanwhile, all the sentences related to three distractors (colored \textit{red} and \textit{orange}) obtain higher gate values, which further help to achieve higher attention weights (highlighted pink in the left part).
Especially, when generating the first distractor, the \textit{red} sentence has the highest attention score, indicating it make the most contributions to the generation.
In summary, the visualization results demonstrate that EDGE provides a good way for the interpretation of the key information of a generated distractor.

\section{Conclusion and Future Work}
In this paper, we propose a novel qu\textbf{E}stion and answer guided \textbf{D}istractor \textbf{GE}neration(EDGE) framework to automatically generate distractors for multiple choice questions in standard English tests.
In EDGE, we design two modules based on attention and gate strategies to reform the passage and question, which then are combined to decode the distractor.
Experimental results on a large-scale public dataset demonstrate the state-of-the-art performance of EDGE and the effectiveness of two reforming modules.

In future work, we will explore two potential directions. 
First, since the beam search ignores the generation diversity, we will explore how to incorporate the prior generated distractor information to guide the generation of successor distractors.
Second, we will work on how to generate the distractors requiring multi-sentence/hop reasoning, which can further improve the plausibility.

\bibliographystyle{coling}
\bibliography{coling2020}

\begin{thebibliography}{}

\bibitem[\protect\citename{Afzal and Mitkov}2014]{Afzal2014AutomaticGO}
Naveed Afzal and Ruslan Mitkov.
\newblock 2014.
\newblock Automatic generation of multiple choice questions using
  dependency-based semantic relations.
\newblock {\em Soft Computing}, 18:1269--1281.

\bibitem[\protect\citename{Araki \bgroup et al.\egroup
  }2016]{Araki2016GeneratingQA}
Jun Araki, Dheeraj Rajagopal, Sreecharan Sankaranarayanan, Susan Holm, Yukari
  Yamakawa, and Teruko Mitamura.
\newblock 2016.
\newblock Generating questions and multiple-choice answers using semantic
  analysis of texts.
\newblock In {\em COLING}.

\bibitem[\protect\citename{Ch and Saha}2018]{Ch2018AutomaticMC}
Dhawaleswar~Rao Ch and Sujan~Kumar Saha.
\newblock 2018.
\newblock Automatic multiple choice question generation from text : A survey.

\bibitem[\protect\citename{Chen \bgroup et al.\egroup }2017]{chen2017neural}
Qian Chen, Xiaodan Zhu, Zhen-Hua Ling, Diana Inkpen, and Si~Wei.
\newblock 2017.
\newblock Neural natural language inference models enhanced with external
  knowledge.
\newblock {\em arXiv preprint arXiv:1711.04289}.

\bibitem[\protect\citename{Chen \bgroup et al.\egroup
  }2019]{Chen2019ReinforcementLB}
Yu~Chen, Lingfei Wu, and Mohammed~J. Zaki.
\newblock 2019.
\newblock Reinforcement learning based graph-to-sequence model for natural
  question generation.
\newblock {\em ArXiv}, abs/1908.04942.

\bibitem[\protect\citename{Devlin \bgroup et al.\egroup
  }2019]{Devlin2019BERTPO}
Jacob Devlin, Ming-Wei Chang, Kenton Lee, and Kristina Toutanova.
\newblock 2019.
\newblock Bert: Pre-training of deep bidirectional transformers for language
  understanding.
\newblock In {\em NAACL-HLT}.

\bibitem[\protect\citename{Du and Cardie}2017]{Du2017IdentifyingWT}
Xinya Du and Claire Cardie.
\newblock 2017.
\newblock Identifying where to focus in reading comprehension for neural
  question generation.
\newblock In {\em EMNLP}.

\bibitem[\protect\citename{Duan \bgroup et al.\egroup
  }2017]{Duan2017QuestionGF}
Nan Duan, Duyu Tang, Peng Chen, and Ming Zhou.
\newblock 2017.
\newblock Question generation for question answering.
\newblock In {\em EMNLP}.

\bibitem[\protect\citename{Gao \bgroup et al.\egroup
  }2018]{Gao2018GeneratingDF}
Yifan Gao, Lidong Bing, Piji Li, Irwin King, and Michael~R. Lyu.
\newblock 2018.
\newblock Generating distractors for reading comprehension questions from real
  examinations.
\newblock In {\em AAAI}.

\bibitem[\protect\citename{Guo \bgroup et al.\egroup
  }2016]{Guo2016QuestimatorGK}
Qi~Guo, Chinmay Kulkarni, Aniket Kittur, Jeffrey~P. Bigham, and Emma Brunskill.
\newblock 2016.
\newblock Questimator: Generating knowledge assessments for arbitrary topics.
\newblock In {\em IJCAI}.

\bibitem[\protect\citename{Ha and Yaneva}2018]{Ha2018AutomaticDS}
Le~An Ha and Victoria Yaneva.
\newblock 2018.
\newblock Automatic distractor suggestion for multiple-choice tests using
  concept embeddings and information retrieval.
\newblock In {\em BEA@NAACL-HLT}.

\bibitem[\protect\citename{Hochreiter and Schmidhuber}1997]{hochreiter1997long}
Sepp Hochreiter and J{\"u}rgen Schmidhuber.
\newblock 1997.
\newblock Long short-term memory.
\newblock {\em Neural computation}, 9(8):1735--1780.

\bibitem[\protect\citename{Kim \bgroup et al.\egroup }2018]{Kim2018ImprovingNQ}
Yanghoon Kim, Hwanhee Lee, Joongbo Shin, and Kyomin Jung.
\newblock 2018.
\newblock Improving neural question generation using answer separation.
\newblock In {\em AAAI}.

\bibitem[\protect\citename{Kumar \bgroup et al.\egroup
  }2015]{kumar2015automatic}
Girish Kumar, Rafael~E Banchs, and Luis~Fernando D'Haro.
\newblock 2015.
\newblock Automatic fill-the-blank question generator for student
  self-assessment.
\newblock In {\em 2015 IEEE Frontiers in Education Conference (FIE)}, pages
  1--3. IEEE.

\bibitem[\protect\citename{Lai \bgroup et al.\egroup }2017]{Lai2017RACELR}
Guokun Lai, Qizhe Xie, Hanxiao Liu, Yiming Yang, and Eduard~H. Hovy.
\newblock 2017.
\newblock Race: Large-scale reading comprehension dataset from examinations.
\newblock In {\em EMNLP}.

\bibitem[\protect\citename{Liang \bgroup et al.\egroup
  }2018]{Liang2018DistractorGF}
Chen Liang, Xiao Yang, Neisarg Dave, Drew Wham, Bart Pursel, and C.~Lee Giles.
\newblock 2018.
\newblock Distractor generation for multiple choice questions using learning to
  rank.
\newblock In {\em BEA@NAACL-HLT}.

\bibitem[\protect\citename{Lin}2004]{Lin2004ROUGEAP}
Chin-Yew Lin.
\newblock 2004.
\newblock Rouge: A package for automatic evaluation of summaries.
\newblock In {\em ACL 2004}.

\bibitem[\protect\citename{Mou \bgroup et al.\egroup }2016]{mou2016natural}
Lili Mou, Rui Men, Ge~Li, Yan Xu, Lu~Zhang, Rui Yan, and Zhi Jin.
\newblock 2016.
\newblock Natural language inference by tree-based convolution and heuristic
  matching.
\newblock In {\em Proceedings of the 54th Annual Meeting of the Association for
  Computational Linguistics (Volume 2: Short Papers)}, volume~2, pages
  130--136.

\bibitem[\protect\citename{Nesterov}1983]{Nesterov1983AMF}
Y.~Nesterov.
\newblock 1983.
\newblock A method for unconstrained convex minimization problem with the rate
  of convergence o(1/$k^2$).

\bibitem[\protect\citename{Papineni \bgroup et al.\egroup
  }2001]{Papineni2001BleuAM}
Kishore Papineni, Salim Roukos, Todd Ward, and Wei-Jing Zhu.
\newblock 2001.
\newblock Bleu: a method for automatic evaluation of machine translation.
\newblock In {\em ACL}.

\bibitem[\protect\citename{Pennington \bgroup et al.\egroup
  }2014]{pennington2014glove}
Jeffrey Pennington, Richard Socher, and Christopher~D. Manning.
\newblock 2014.
\newblock Glove: Global vectors for word representation.
\newblock In {\em Empirical Methods in Natural Language Processing (EMNLP)},
  pages 1532--1543.

\bibitem[\protect\citename{Qiu \bgroup et al.\egroup
  }2019]{DBLP:conf/cikm/QiuW019}
Zhaopeng Qiu, Xian Wu, and Wei Fan.
\newblock 2019.
\newblock Question difficulty prediction for multiple choice problems in
  medical exams.
\newblock In {\em {CIKM}}, pages 139--148. {ACM}.

\bibitem[\protect\citename{Sakaguchi \bgroup et al.\egroup
  }2013]{Sakaguchi2013DiscriminativeAT}
Keisuke Sakaguchi, Yuki Arase, and Mamoru Komachi.
\newblock 2013.
\newblock Discriminative approach to fill-in-the-blank quiz generation for
  language learners.
\newblock In {\em ACL}.

\bibitem[\protect\citename{Stasaski and Hearst}2017]{Stasaski2017MultipleCQ}
Katherine Stasaski and Marti~A. Hearst.
\newblock 2017.
\newblock Multiple choice question generation utilizing an ontology.
\newblock In {\em BEA@EMNLP}.

\bibitem[\protect\citename{Vaswani \bgroup et al.\egroup
  }2017]{Vaswani2017AttentionIA}
Ashish Vaswani, Noam Shazeer, Niki Parmar, Jakob Uszkoreit, Llion Jones,
  Aidan~N. Gomez, Lukasz Kaiser, and Illia Polosukhin.
\newblock 2017.
\newblock Attention is all you need.
\newblock In {\em NIPS}.

\bibitem[\protect\citename{Yang \bgroup et al.\egroup
  }2017]{Yang2017SemiSupervisedQW}
Zhilin Yang, Junjie Hu, Ruslan Salakhutdinov, and William~W. Cohen.
\newblock 2017.
\newblock Semi-supervised qa with generative domain-adaptive nets.
\newblock In {\em ACL}.

\bibitem[\protect\citename{Yang \bgroup et al.\egroup }2019]{Yang2019XLNetGA}
Zhilin Yang, Zihang Dai, Yiming Yang, Jaime~G. Carbonell, Ruslan Salakhutdinov,
  and Quoc~V. Le.
\newblock 2019.
\newblock Xlnet: Generalized autoregressive pretraining for language
  understanding.
\newblock {\em ArXiv}, abs/1906.08237.

\bibitem[\protect\citename{Zhou \bgroup et al.\egroup }2017]{Zhou2017NeuralQG}
Qingyu Zhou, Nan Yang, Furu Wei, Chuanqi Tan, Hangbo Bao, and Ming Zhou.
\newblock 2017.
\newblock Neural question generation from text: A preliminary study.
\newblock In {\em NLPCC}.

\bibitem[\protect\citename{Zhou \bgroup et al.\egroup
  }2019]{zhou2019coattention}
Xiaorui Zhou, Senlin Luo, and Yunfang Wu.
\newblock 2019.
\newblock Co-attention hierarchical network: Generating coherent long
  distractors for reading comprehension.

\end{thebibliography}

\end{document}